\documentclass[sigconf, 10pt]{acmart}

\AtBeginDocument{%
  }

\setcopyright{rightsretained}
\copyrightyear{2025}
\acmYear{2025}
\acmDOI{XXXXXXX.XXXXXXX}

\acmConference[{ACM ICAIF '25}]%
  {The 2nd Workshop on LLMs and Generative AI for Finance, November 15, Singapore}%
  {November 15}%
  {Singapore}

\usepackage{algorithm}
\usepackage{algpseudocode}
\usepackage{booktabs}
\usepackage{longtable}

\begin{document}

 \title{CompactPrompt: A Unified Pipeline for Prompt and Data Compression in LLM Workflows}

\author{Joong Ho Choi}
\email{joongho.choi@bny.com}
\affiliation{%
  \institution{BNY}
  \city{Pittsburgh}
  \state{PA}
  \country{USA}
}

\author{Jiayang Zhao}
\email{jiayang.zhao@bny.com}
\affiliation{%
  \institution{BNY}
  \city{Pittsburgh}
  \state{PA}
  \country{USA}
}

\author{Jeel Shah}
\email{jeel.shah@bny.com}
\affiliation{%
  \institution{BNY}
  \city{Pittsburgh}
  \state{PA}
  \country{USA}
}

\author{Ritvika Sonawane}
\email{rsonawan@andrew.cmu.edu}
\affiliation{%
  \institution{Carnegie Mellon University}
  \city{Pittsburgh}
  \state{PA}
  \country{USA}
}
\authornote{Project completed during internship at BNY}

\author{Vedant Singh}
\email{vedant.singh@bny.com}
\affiliation{%
  \institution{BNY}
  \city{Pittsburgh}
  \state{PA}
  \country{USA}
}

\author{Avani Appalla}
\email{avani.appalla@bny.com}
\affiliation{%
  \institution{BNY}
  \city{Pittsburgh}
  \state{PA}
  \country{USA}
}

\author{Will Flanagan}
\email{william.flanagan@bny.com}
\affiliation{%
  \institution{BNY}
  \city{Pittsburgh}
  \state{PA}
  \country{USA}
}

\author{Filipe Condessa}
\email{filipe.condessa@bny.com}
\affiliation{%
  \institution{BNY}
  \city{Pittsburgh}
  \state{PA}
  \country{USA}
}

\begin{abstract}
Large Language Models (LLMs) deliver powerful reasoning and generation capabilities but incur substantial run-time costs when operating in agentic workflows that chain together lengthy prompts and process rich data streams. We introduce CompactPrompt, an end-to-end pipeline that merges hard prompt compression with lightweight file-level data compression. CompactPrompt first prunes low-information tokens from prompts using self-information scoring and dependency-based phrase grouping. In parallel, it applies n-gram abbreviation to recurrent textual patterns in attached documents and uniform quantization to numerical columns, yielding compact yet semantically faithful representations. Integrated into standard LLM agents, CompactPrompt reduces total token usage and inference cost by up to 60\% on benchmark dataset like TAT-QA \cite{zhu-etal-2021-tat} and FinQA \cite{chen2021finqa}, while preserving output quality (Results in less than 5\% accuracy drop for Claude-3.5-Sonnet \cite{anthropic_claude3_5_sonnet}, and GPT-4.1-Mini \cite{openai_chatgpt4.1_mini})
CompactPrompt helps visualize real-time compression decisions and quantify cost–performance trade-offs, laying the groundwork for leaner generative AI pipelines. Demo is available at \newline
\href{https://www.youtube.com/watch?v=xr_7Wbzemzg}{\text{https://www.youtube.com/watch?v=xr\_7Wbzemzg}}
\end{abstract}



\renewcommand{\shortauthors}{BNY et al.}
\maketitle

\section{Introduction}
Global financial institutions were among the first to integrate large-scale generative models into production systems, driven by the need for both operational efficiency and consistent output quality. Prompt compression—the process of algorithmically removing or rewriting superfluous tokens from user inputs—directly reduces computational overhead (fewer tokens \(\rightarrow\) lower inference cost and latency) and can standardize prompts across users with varying levels of expertise. Prior work has shown that verbose or unstructured inputs not only increase model cost but also introduce variability in generated outputs \cite{brown2020language, shi2023optimizing}. By applying prompt-level optimization (\emph{e.g.}, token pruning, heuristic rewriting, or LLM-based summarization), institutions can achieve measurable cost savings while simultaneously improving response consistency and reliability. This dual impact makes prompt compression a compelling technique for enterprise-scale LLM applications.

Empirical studies on Large language models \cite{levy2024tasktokensimpactinput}, \cite{li2024longcontextllmsstrugglelong} reveal that beyond a certain length, adding tokens degrades reasoning performance. In \cite{levy-etal-2024-task}, a marked decline in various reasoning benchmarks once prompts approach roughly 3,000 tokens, which are well below the context-window limits of contemporary transformer models. 

One manifestation of this limitation is the “lost-in-the-middle” effect \cite{liu2023lostmiddlelanguagemodels}: as input sequences lengthen, transformer attention mechanisms tend to over-emphasize tokens at the beginning and end of the prompt, allocating disproportionately less weight to information appearing midway through. The result is an impaired ability to integrate all relevant context during multi-step reasoning tasks. 

Together, these findings highlight a fundamental constraint in current transformer architectures: their capacity to reason coherently over very long sequences is inherently bounded. Although prompt-engineering techniques, such as CoT, can partially mitigate this effect, they cannot fully overcome it. Practitioners must therefore strike a careful balance; they need to provide sufficient context to support complex reasoning without exceeding the model’s effective working window, and this is where prompt compression comes in.

To achieve the previously mentioned balance, two principal classes of solutions for prompt compression \cite{li2024promptcompressionlargelanguage}. Hard prompt compression techniques, including SelectiveContext \cite{selectivecontext} and LLMLingua \cite{llmlingua}. Soft prompt compression methods such as GIST \cite{gist} and 500$\times$-Compressor \cite{500xcompressor}. 
Crucially, all existing work operates on the prompt in isolation, ignoring another major contributor to context size: external data attachments. Traditional file compression (\emph{e.g.}, gzip, numeric quantizers) reduces storage and transmission costs but is incompatible with token-based LLM APIs. 
In this paper, we introduce `CompactPrompt', an integrated framework for simultaneous prompt and file-level compression in LLM pipelines that address these gaps by introducing context compression. It is an end-to-end, training-free pipeline that unifies:
\begin{itemize}
    \item Hard Prompt Pruning via token-level self-information scoring and dependency-driven phrase grouping.
 
    \item Textual n-gram Abbreviation, which constructs a reversible dictionary of high-frequency multi-word expressions in attached documents and replaces them with unique, short, human-readable tokens.
 
    \item Numeric Quantization, which maps floating-point columns to fixed-bit integer codes under user-specified error tolerances, preserving analytical fidelity.
\end{itemize}

\section{Related Work}

SelectiveContext \cite{selectivecontext}, a pioneering prompt compression method, enables LLMs to process 2x more content while saving 36\% memory and 32\% inference time using vicuna-13b\cite{lmsys2024vicuna13b-v1.5} in summary generation task. It uses self-information to identify and delete less informative parts of prompts and employs Spacy's syntactic parsing to ensure grammatical integrity.

LLMLingua \cite{llmlingua} achieves up to 20x prompt compression with minimal performance loss using trained language models. It employs a coarse-to-fine compression method, featuring a budget controller for semantic integrity. 

Another method is Nano-Capsulator \cite{chuang2024learningcompresspromptnatural}, which transforms an input prompt into a succinct natural-language summary, which is then fed into the target LLM. By pruning irrelevant content and reconstructing fluent sentences, its fine-tuned Vicuna-7B compressor \cite{lmsys2023vicuna7b-v1.5}
 runs independently of the downstream model. Unlike off-the-shelf summarizers, Nano-Capsulator incorporates a semantic-preservation loss to retain task-critical meaning and a reward function to optimize prompt utility, boosting end-task performance. On the downside, the extra compression network incurs memory overhead, and its two-stage inference pipeline demands more compute than simple encoding.

Beyond these, two other general “hard-prompt” compressors have emerged. LongLLMLingua extends LLMLingua by reordering documents and recovering subsequences to double the effective window \cite{jiang2024longllmlinguaacceleratingenhancingllms}, while AdaComp adaptively selects context based on query complexity and retrieval quality \cite{zhang2024adacompextractivecontextcompression}. LLMLingua-2 further refines this via data-distillation to build a compressed corpus and a classifier that predicts which tokens to keep \cite{pan2024llmlingua2datadistillationefficient}. 

Although prior work (\emph{e.g.} \cite{zhang2025empiricalstudypromptcompression}, \cite{li2024promptcompressionlargelanguage}) has studied a variety of hard prompt compression and soft prompt rewriting strategies, we found no existing method that directly applies n-gram abbreviation in the context of prompt pruning and data compression. Most known approaches either operate at the token-level removal (filtering \cite{llmlingua}), sentence-level selection, or generative rewriting (SCOPE \cite{zhang2025scopegenerativeapproachllm}). 

\section{Design}

Our combination of hard prompt pruning, n-gram abbreviation and numeric quantization is novel and addresses a gap in the literature by allowing interpretable, reversible compression that leverages repeated multi-token patterns rather than purely individual token importance.

Hard prompt pruning targets redundant or low-information content in natural language prompts, preserving semantic meaning while reducing token count; our hard prompt compression strategy leverages both static
and dynamic self-information measures to identify and prune
low-value content from the input prompt, ensuring that
the most informative tokens are preserved within a user-
specified budget. N-gram abbreviation complements this by focusing on repetitive patterns in attached documents, offering lossless compression for domain-specific terminology or common phrases. Numeric quantization addresses the often-overlooked issue of floating-point precision in data-heavy contexts, allowing for significant reduction in token usage for numerical data while maintaining analytical fidelity within user-defined error bounds. This holistic approach ensures that CompactPrompt can effectively compress diverse input types commonly found in real-world LLM applications, from text-heavy prompts to data-rich attachments, without requiring model retraining or sacrificing human readability.

\subsection{Hard Prompt Compression}

\subsubsection{Static Self-Information}
We begin by constructing a large offline corpus-comprising sources, consisting of Wikipedia, ShareGPT conversations, and arXiv articles. Then we compute unigram frequencies f(t) for every token t as shown below.

\begin{align}
f(t)
&= \frac{1}{N}\sum_{i=1}^{N}\mathbf{1}\{w_i = t\} \label{eq:freq1} = \frac{\#\{\,i : w_i = t\}}{N}
\end{align}

From these frequencies, we derive the static self-information score using $[I(T) = -\log_2 p(T)]$.
\subsubsection{Dynamic Self-Information}
To capture context-sensitive importance, we query a pretrained LLM or a lightweight scoring agent at runtime. Given a token t and its preceding context (c), we obtain the model probability ($P_{\text{model}}(t \mid c)$) and compute dynamic self-information. This score reflects the novel information contributed by t in its specific prompt context, complementing the corpus-level view.

\subsubsection{Combined Scoring and Phrase-Level Pruning}

We integrate static and dynamic measures through a weighted combination strategy, employing a simple arithmetic mean when both scores exhibit less than 10\% relative difference, as shown below in formula ~\eqref{eq:delta} and~\eqref{eq:comb}. 


\begin{align}
\Delta &=
\frac{\bigl\lvert s_{\mathrm{dyn}} - s_{\mathrm{stat}}\bigr\rvert}{s_{\mathrm{stat}}}
\label{eq:delta}\\
C\bigl(s_{\mathrm{stat}},s_{\mathrm{dyn}}\bigr)
&=
\begin{cases}
\displaystyle
\frac{s_{\mathrm{stat}} + s_{\mathrm{dyn}}}{2}, & \Delta \le 0.1,\\[8pt]
s_{\mathrm{dyn}}, & \Delta > 0.1.
\end{cases}
\label{eq:comb}
\end{align}

The reasoning behind 10\% threshold is because lower (\emph{e.g.} 5\%) or higher (\emph{e.g.} 15\%) cut-offs tilt too far toward one method, either over-favoring dynamic noise or under-reacting to true context changes. Ten percent offers a pragmatic compromise. 

When both methods produce similar scores, their average provides a stable estimate that benefits from both corpus-level statistics and contextual information. However, when scores differ significantly, the dynamic method's ability to capture context-sensitive importance through querying the pre-trained model at runtime is more reliable than static corpus statistics. 
\subsection{Textual n-gram Abbreviation}
N-gram abbreviation uses a method similar to LZW (Lempel–Ziv–Welch) algorithm \cite{welch1984technique} that compresses text by losslessly replacing common sequences with short, human-readable tokens, so you can search, index or apply Natural Language Processing (NLP) directly on the stream without full decompression. It delivers substantial size reduction with simple table-lookup encoding-more transparent and lightweight than heavy binary compression like ZIP\cite{pkware1989zip} or bzip2\cite{seward1996bzip2}, and reversible (unlike lossy embeddings) at minimal compute cost.
To compress attached text, we employ a user-configurable n-gram abbreviation pipeline. 

\subsubsection{Extraction and Frequency Analysis}
First, the user specifies an n-gram length (n) (commonly between 2 and 5). We extract all n-grams from the document corpus and compute their frequencies. These frequencies are visualized in a histogram, enabling the identification of the top (K) most frequent patterns (typically (K=100 to 150)).

\subsubsection{Dictionary Construction}
We assign each of the top-(K) n-grams a unique placeholder token (\emph{e.g.}, “ABC1”, “BD2”). This mapping is stored in a metadata table, ensuring lossless round-trip reconstruction.

\subsubsection{Contextual Replacement and Reversibility}
We replace every occurrence of the most-common n-grams with their placeholders, taking care to resolve overlapping matches and preserve punctuation. At downstream stages or upon user request, placeholders are substituted back to their original n-grams using the metadata table, guaranteeing exact reversibility.

\subsection{Numerical Quantization}
For numerical columns in structured data attachments, we implement two complementary quantization schemes to reduce token payload while bounding approximation error, namely uniform integer quantization and K-means based quantization.

\subsubsection{Uniform Integer Quantization}
Given a numeric column (x), we compute its minimum and maximum values $(\min_x, \max_x)$ and select a bit-width (b), yielding $(L=2^b)$ quantization levels. Each value ($x_i$) is encoded as shown in equation \eqref{eq:quant} below and then stored as an integer.

\begin{equation}
q_i \;=\; 
\mathrm{round}\!\left(\frac{x_i - \min_x}{\max_x - \min_x}\,(L - 1)\right)
\label{eq:quant}
\end{equation}

The tuple $(\min_x,\max_x,b)$ is retained to reconstruct ($\hat x_i$) with maximum absolute error as defined in equation \eqref{eq:optimus} below.

\begin{equation}
\begin{aligned}
\hat x_i &= \min_x \;+\;\frac{q_i}{L-1}\,\bigl(\max_x - \min_x\bigr),\\
\varepsilon_{\max} &= \frac{\max_x - \min_x}{L-1}\,
\end{aligned}
\label{eq:optimus}
\end{equation}

\subsubsection{K-Means-Based Quantization}
Alternatively, we apply k-means clustering to the column values, choosing (k) centroids $(\mu_1,\dots,\mu_k)$. Each $x_i$ is mapped to the nearest centroid index. Reconstruction uses the stored centroids, which minimizes average squared error.

\subsection{Representative Example Selection for Few-Shot}
To further reduce contextual size when including few-shot exemplars, we select a small set (i.e. 3) of representative data points via clustering and silhouette analysis.

\subsubsection{Embedding and Normalization}
We choose all-mpnet-base-v2 \cite{all-mpnet-base-v2} for the embedding model. We embedded textual data into high-dimensional vectors and standardized numeric features (zero mean, unit variance).

\subsubsection{Clustering with Silhouette Optimization}
We run k-means over the embedded or normalized data for $k\in{5-50}$. For each (k), we compute the average silhouette score—a measure of cluster cohesion versus separation—to identify the optimal cluster count $(k^*)$, given the value of k that maximizes this score.

\subsubsection{Representative Points Selection}
Within each of the $(k^*)$ clusters, we select the point closest to the centroid as a “prototype.” These prototypes serve as few-shot examples, capturing the diversity of the dataset in a compact set.

\subsection{Semantic Fidelity and Similarity Metrics}
To evaluate the semantic integrity of compressed outputs, we employ embedding-based similarity measures.

\subsubsection{Full-Dimensional Cosine Similarity}
Original and compressed text segments are encoded with all-mpnet-base-v2 \cite{all-mpnet-base-v2}. We compute
$\mathrm{cosine}(E_{\text{orig}}, $ $E_{\text{comp}})$ and report both the mean and 5th percentile scores to ensure worst-case fidelity remains high.

\subsubsection{Human Validation}
The scale of evaluation ranges from 1 to 5, with 5 being completely identical and thus good. Three evaluators were trained to have same evaluation standard. With 15 examples from each TAT-QA and Fin-QA(overall 30), they were shown randomly sampled examples that represent different scores asked to evaluate other 15 examples from each TAT-QA and Fin-QA(overall 30 again) to ensure their score alignment. 

After this training, they rated 90 original \& compressed prompt pairs from TAT-QA and Fin-QA for semantic equivalence. We set the score ranging from 1-5, with 1 being completely different and 5 being completely identical. The mean score (4.1/5) correlated with cosine similarities. Fewer than 5 percents of cases showed high vector similarity but lower human ratings, highlighting rare nuance shifts missed by embedding alone.

\section{CompactPrompt Tool}

\subsection{CompactPrompt GUI Workflow}

\begin{figure}[h]
  \centering
  \includegraphics[width=0.5\textwidth]{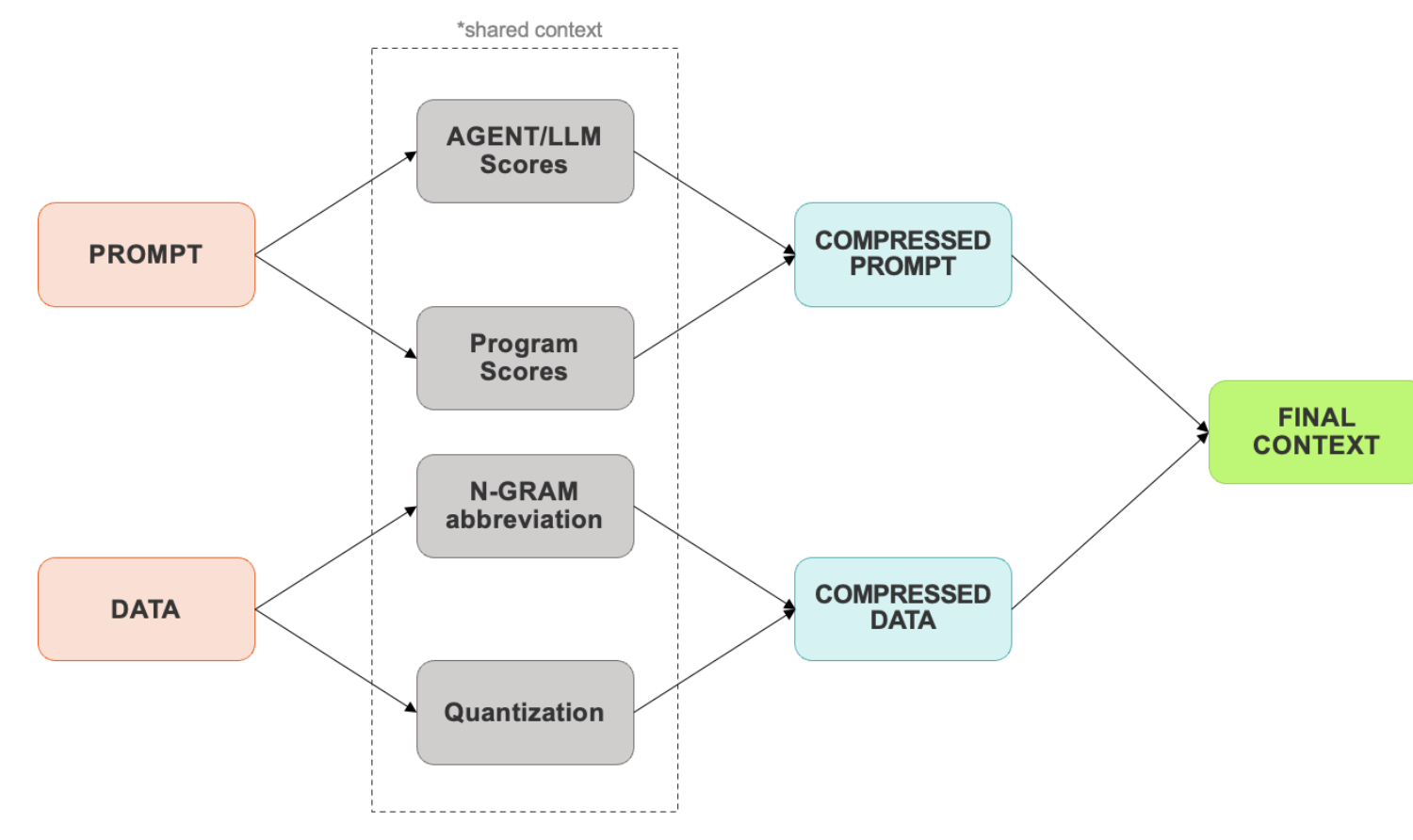}
  \caption{CompactPrompt Workflow}
  \label{fig:workflow}
\end{figure}

The Tool is divided into two parts: “Prompt Compression” and “Data Compression.” Figure \ref{fig:workflow} shows the workflow of the tool.
In the Prompt Compression view shown in Figure 2, users can paste or type their full prompt in the text editor and can choose which LLM calculates self-information scores via a Scorer agent. At the moment, we support GPT-4-Omni \cite{openai_chatgpt4o_omni}, GPT-4.1-Mini \cite{openai_chatgpt4.1_mini}, Claude 3.5 Sonnet \cite{anthropic_claude3_5_sonnet}, and Llama-3.3-70B-instruct \cite{llama_3.3}.

\begin{figure}[h]
  \centering
  \includegraphics[width=0.5\textwidth]{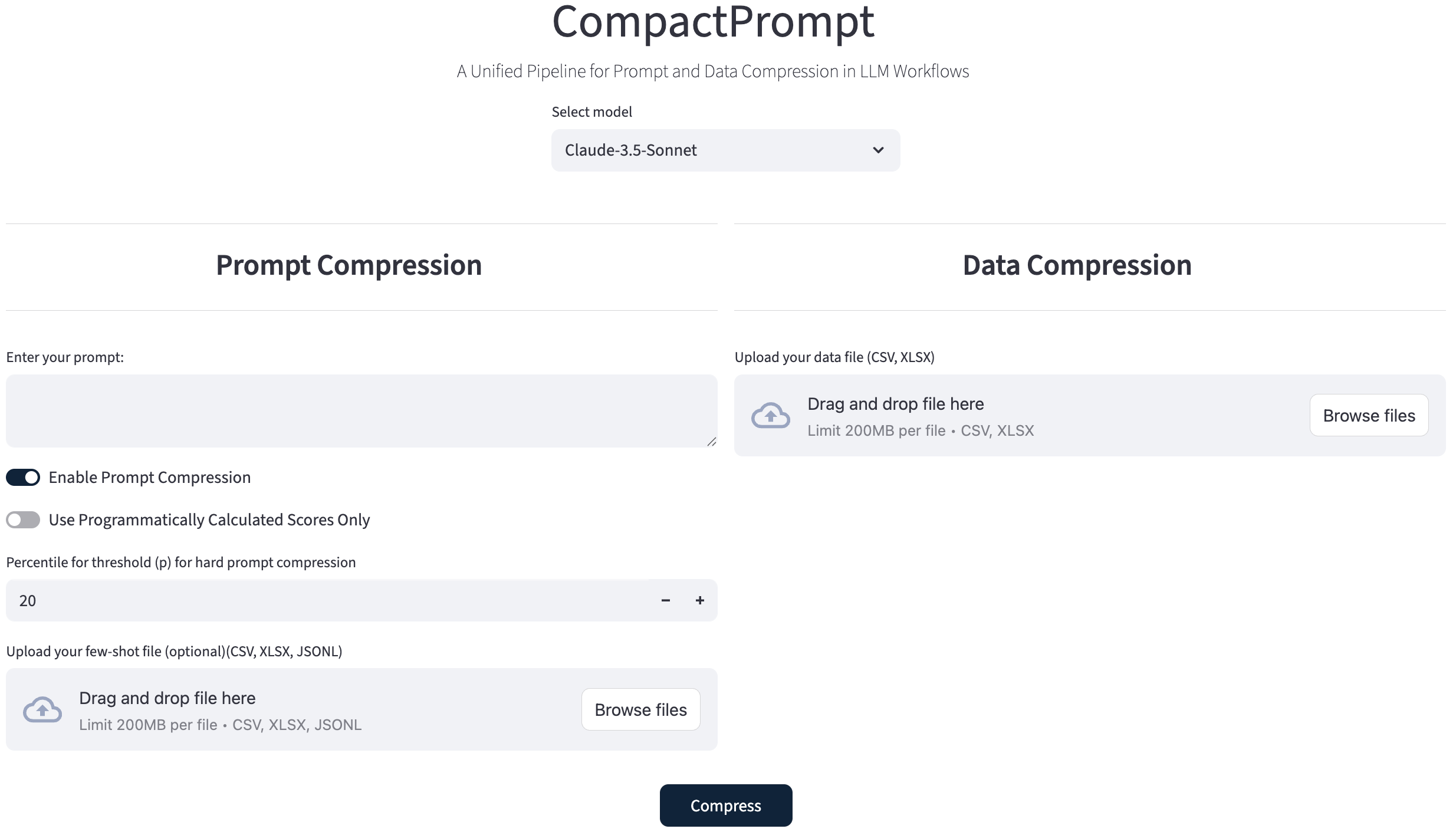}
  \caption{CompactPrompt App Preview}
  \label{fig:finalapp}
\end{figure}

Refer to the Demo video:
\newline
\href{https://www.youtube.com/watch?v=xr_7Wbzemzg}{\text{https://www.youtube.com/watch?v=xr\_7Wbzemzg}}

\subsection{CompactPrompt non-GUI/Library version Workflow}
Following the development of CompactPrompt as a GUI application, we recognized the need to extend its utility beyond individual prompt engineering workflows to accommodate enterprise-scale deployment and systematic integration within organizational AI infrastructure. To address this requirement, we architected a library implementation through the UnifiedCompactPromptPipeline (UCPP) class, which encapsulates our comprehensive compression methodology within a programmatically accessible framework. This library instantiation enables seamless integration with existing financial technology stacks, allowing compression capabilities to be incorporated directly into automated trading systems, risk management platforms, and regulatory compliance pipelines. The UCPP implementation maintains the full algorithmic rigor of our original approach while providing enterprise-grade scalability and operational flexibility essential for production financial environments.
\begin{figure}[h]
  \centering
  \includegraphics[width=0.5\textwidth]{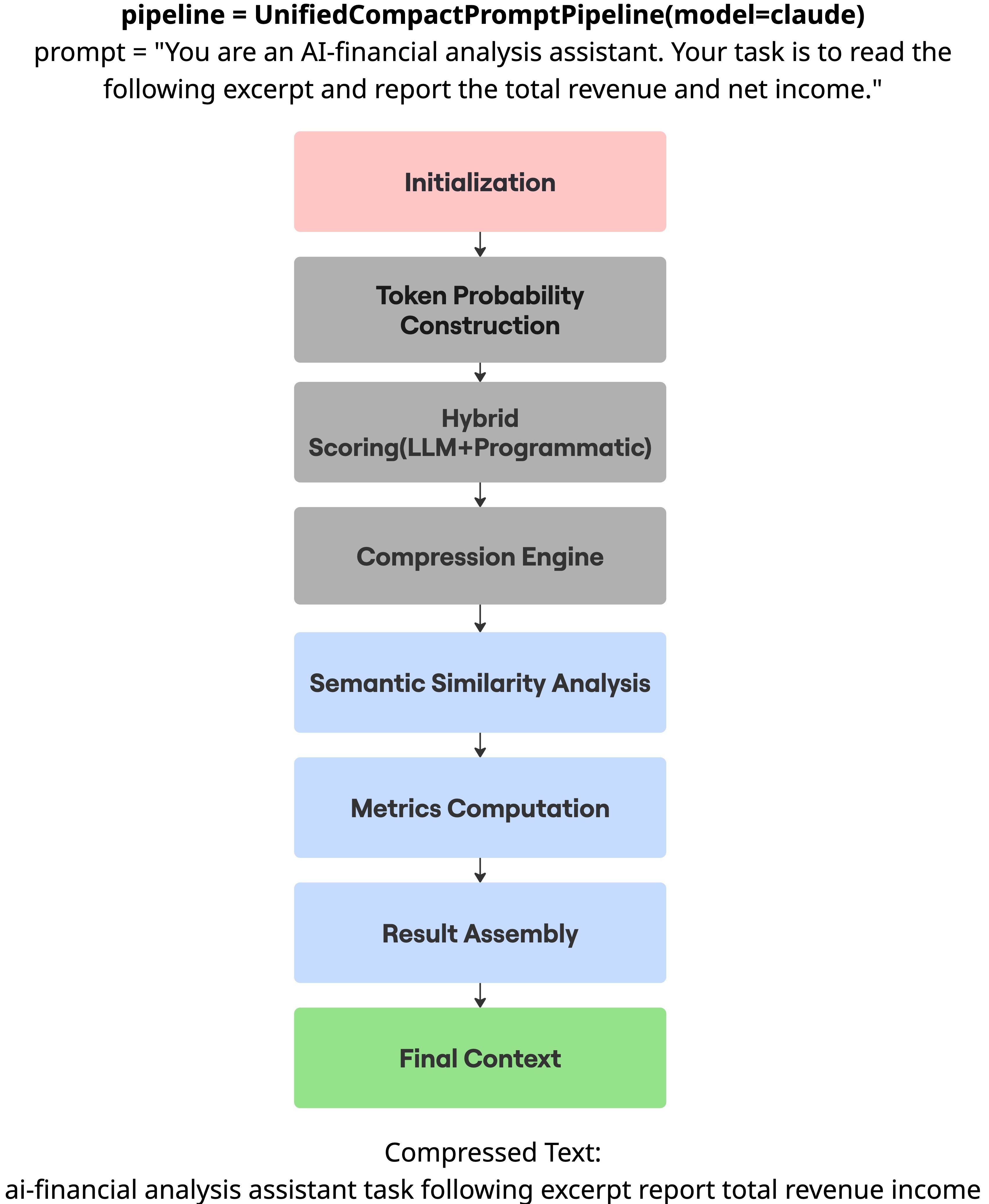}
  \caption{Non-GUI Preview}
  \label{fig:non-gui}
\end{figure}

The UnifiedCompactPromptPipeline orchestrates an eight-stage workflow encompassing initialization, token probability construction, hybrid scoring through combined large language model and programmatic evaluation, our compression engine, semantic similarity analysis, comprehensive metrics computation, result assembly, and utility function execution, as shown in Figure 3. Upon instantiation with model-specific parameters \newline (\emph{e.g.}, $pipeline = \texttt{UCPP(model='Claude-3.5-Sonnet')}$), the system processes input prompts through this deterministic pipeline, ultimately delivering both the  compressed prompt and detailed compression analytics directly to the terminal interface. This programmatic approach fundamentally transforms CompactPrompt from a standalone tool into a deployable enterprise asset, enabling organizations to systematically reduce LLM operational costs across their entire AI infrastructure while maintaining quantifiable performance standards. The library's terminal-based output facilitates direct integration with database query systems and automated trading algorithms, thereby establishing a seamless bridge between prompt optimization and production financial applications where computational efficiency directly translates to competitive advantage and operational cost reduction.
\newline
The UnifiedCompactPromptPipeline orchestrates a meticulously engineered eight-stage workflow encompassing initialization, token probability construction, hybrid scoring through combined large language model and programmatic evaluation, our compression engine, semantic similarity analysis, comprehensive metrics computation, result assembly, and utility function execution. Upon instantiation with model-specific parameters 
\newline (\emph{e.g.}, $pipeline = \texttt{UCPP(model='Claude-3.5-Sonnet')}$), 
the system processes input prompts through this deterministic pipeline, ultimately delivering both the optimally compressed prompt and detailed compression analytics directly to the terminal interface. This programmatic approach fundamentally transforms CompactPrompt from a standalone tool into a deployable enterprise asset, enabling organizations to systematically reduce LLM operational costs across their entire AI infrastructure while maintaining quantifiable performance standards. The library's terminal-based output facilitates direct integration with database query systems and automated trading algorithms, thereby establishing a seamless bridge between prompt optimization and production financial applications where computational efficiency directly translates to competitive advantage and operational cost reduction.

\section{Downstream Task Performance}
\subsection{Task Setup} 
To evaluate our compression methodology, we conduct experiments on two complementary datasets: TAT-QA \cite{zhu-etal-2021-tat} and Fin-QA \cite{chen2021finqa}. TAT-QA is a large-scale benchmark for question answering that presents a challenge requiring integration of structured tabular data with unstructured narrative passages, where queries demand cross-referential reasoning between financial tables and their explanatory annotations. The dataset's long-form contexts populated with redundant tabular tokens create natural vulnerabilities to token bloat and repetitive patterns which are precisely the inefficiencies our CompactPrompt methodology addresses through phrase grouping and abbreviation. Fin-QA is derived from financial reports and emphasizes quantitative reasoning operations while maintaining semantic grounding in specialized financial discourse. Its stringent requirements for numerical precision and frequent deployment of recurring financial terminology generate verbose contexts that interweave structured numerical data with dense financial narratives, reflecting high-stakes applications where errors have profound consequences.
 
The strategic pairing of these datasets creates a comprehensive evaluation framework spanning multiple complexity dimensions. Structurally, TAT-QA's hybrid table-narrative format contrasts with Fin-QA's domain-dense financial text, collectively covering heterogeneous input modalities from structured tabular data to numerically intensive content.  While TAT-QA emphasizes cross-referential information synthesis, Fin-QA prioritizes precision arithmetic and financial logic, ensuring our compression methodology is evaluated on both semantic preservation and quantitative accuracy retention. The datasets exhibit complementary error profiles; while TAT-QA failures typically manifest as information retrieval breakdowns, Fin-QA errors arise from numerical miscalculations or semantic misinterpretation of financial terminology, stress-testing our approach across information compression fidelity and numerical compression safety. This combination provides rigorous validation across the full spectrum of contemporary QA challenges.

In addition to dataset-level evaluation, we conduct a controlled sensitivity study on the hyper-parameters governing our abbreviation-based data compression, specifically the choice of n in n-gram extraction and the number of top-n frequent tokenized sequences retained. By systematically varying these parameters, we assess how compression aggressiveness influences downstream accuracy across both TAT-QA and Fin-QA. The smaller configurations such as bi-grams with Top-3 yield modest reductions (1.12×) compared with baseline prompt, while larger n-grams combined with broader Top-n lists produce progressively stronger effects, with 4-grams at Top-5 achieving the highest compression ratio of 1.44×. This trend illustrates the inherent trade-off between efficiency and fidelity—higher n and Top-n settings maximize token savings but risk discarding subtle semantic cues. To ensure generalizability, the evaluation spans a diverse set of state-of-the-art LLMs, including GPT-4-Omni \cite{openai_chatgpt4o_omni}, GPT-4.1-Mini \cite{openai_chatgpt4.1_mini}, Claude 3.5 Sonnet \cite{anthropic_claude3_5_sonnet}, and Llama-3.3-70B-instruct \cite{llama_3.3}, thereby capturing model-specific sensitivities to context abbreviation. The results provide actionable insights into hyper-parameter tuning, guiding practitioners to balance efficiency and accuracy in real-world deployments.

\subsection{Results}

We evaluated CompactPrompt across TAT-QA and Fin-QA, comparing multiple compression strategies and ablations over n-gram length and top-n frequency thresholds, as shown in the tables in the appendix. Experiments span four leading GPT-4-Omni \cite{openai_chatgpt4o_omni}, GPT-4.1-Mini \cite{openai_chatgpt4.1_mini}, Claude 3.5 Sonnet \cite{anthropic_claude3_5_sonnet}, and Llama-3.3-70B-instruct \cite{llama_3.3}. And we found diverse sensitivities on the models to compression and abbreviation.

\subsubsection{\textbf{Ablation on N-gram and Top-N}\\}
\textbf{Top-N (T)} controls the breadth of the abbreviation dictionary. Small T focuses on highly redundant patterns (maximizing token savings per substitution) and minimizes disruption; large T quickly enters diminishing returns and elevates semantic risk.

\textbf{N-gram size (G)} governs the semantic granularity of each substitution. In practice, we target short, highly recurrent bi-grams (\emph{e.g.} “per share,” “interest expense”) because they yield the biggest compression gains with zero ambiguity. Very long n-grams often encode one-off, document-specific phrasing—but our scheme remains lossless by design: every replaced n-gram is recorded in a reversible lookup table, so the exact original text can be reconstructed. In other words, even if we substitute long, instance-specific sequences, no information is ever discarded—only temporarily aliased and fully recoverable. 

\begin{figure}[h]
  \centering
  \includegraphics[width=0.5\textwidth]{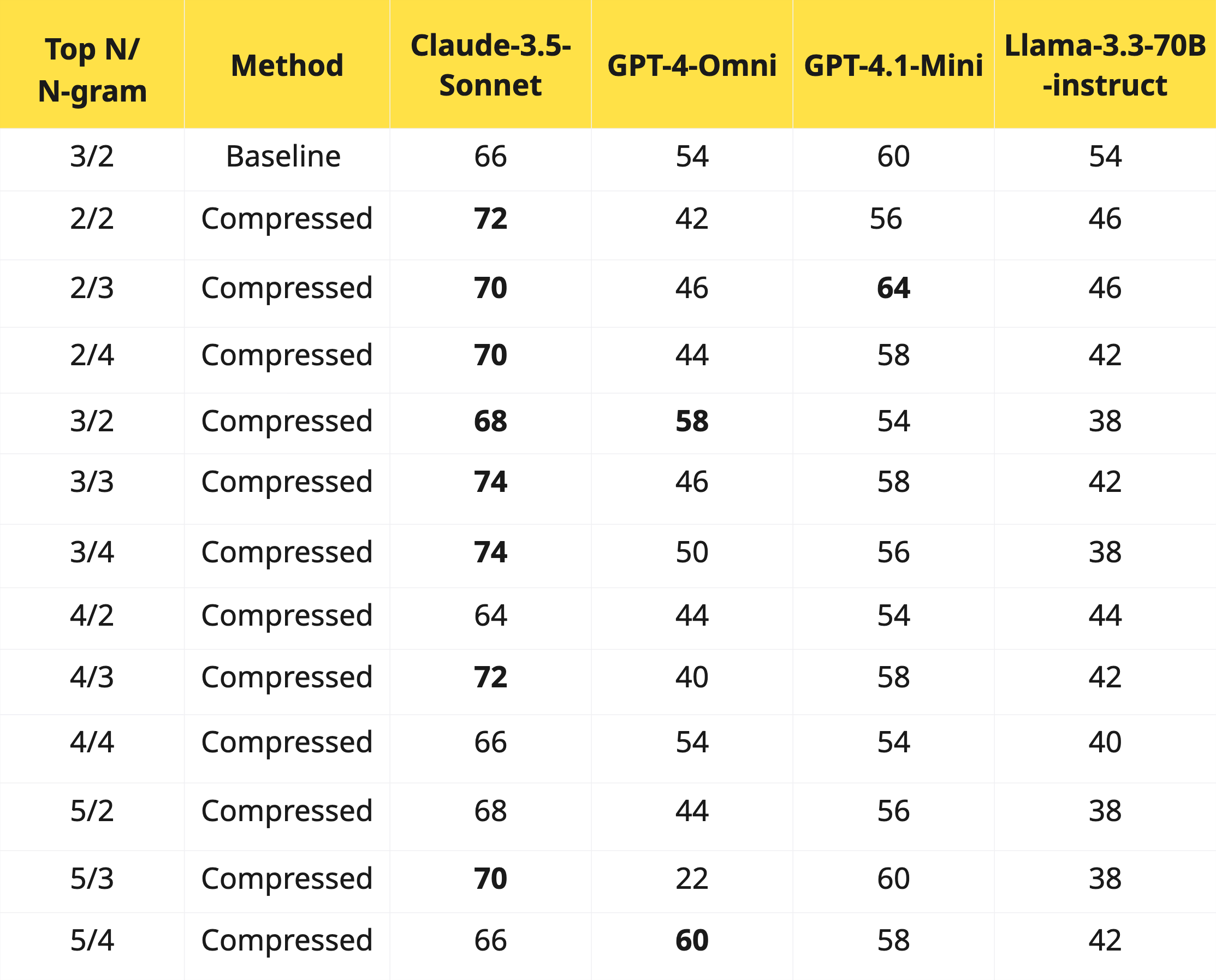}
  \caption{Table showing performance of different models across varying Top N and N-gram configurations}
  \label{fig:heattatqa}
\end{figure}



\textbf{Best configuration: }
 The largest improvement occurs at $T=3$,$G=2$ ($\Delta = +5.0$; annotated “Top 1” in the figure). Targeting only the three most frequent bi-grams concentrates compression on truly repetitive phrasing while avoiding substitutions that would disrupt local semantics. This setting delivers a strong cost–performance trade-off across models. 

\textbf{Next-best configurations: }
Moderate gains are observed for $T=2, G=4$ ($\Delta \approx +1.0$; “Top 2”) and $T=3, G=4$ ($\Delta \approx +0.5$; “Top 3”). These combinations expand coverage to slightly longer, domain-specific expressions (\emph{e.g.}, recurring financial terms) without overwhelming the prompt with substitutions. 

\textbf{Failure region:}
Performance drops cluster in the aggressive quadrant $T\geq4$ (across most G), where crosses dominate. Abbreviating too many patterns increases the chance of (i) replacing context-bearing phrases, (ii) creating dense placeholder text that is harder for models to resolve, and (iii) introducing overlap/collision effects between placeholders—each of which degrades reasoning fidelity. 

\textbf{Model sensitivity:}
Claude 3.5 Sonnet and GPT-4.1-Mini show consistent gains in the conservative regime $T\leq3$with $G \in \{2,4\}$. GPT-4-Omni exhibits mixed responses (both sizable gains and losses), suggesting tighter guardrails on $T$. Llama-3.3-70B is comparatively fragile: it benefits at $G=2, T\leq3$ but degrades rapidly as $T$ increases. 

\subsubsection{\textbf{CompactPrompt Performance with best performing N-gram and TopN values}\\}
After applying CompactPrompt with n-gram = 2 and TopN = 3, the size of the prompts for both TAT-QA and Fin-QA is cut roughly in half. TAT-QA’s token count achieves about a 2.35× reduction, or 58\% fewer tokens, and Fin-QA’s about a 2.12× reduction, or 53\% fewer tokens. This token reduction is achieved while generally maintaining similar or slight improvement for the models.

\begin{figure}[h]
  \centering
  \includegraphics[width=0.5\textwidth]{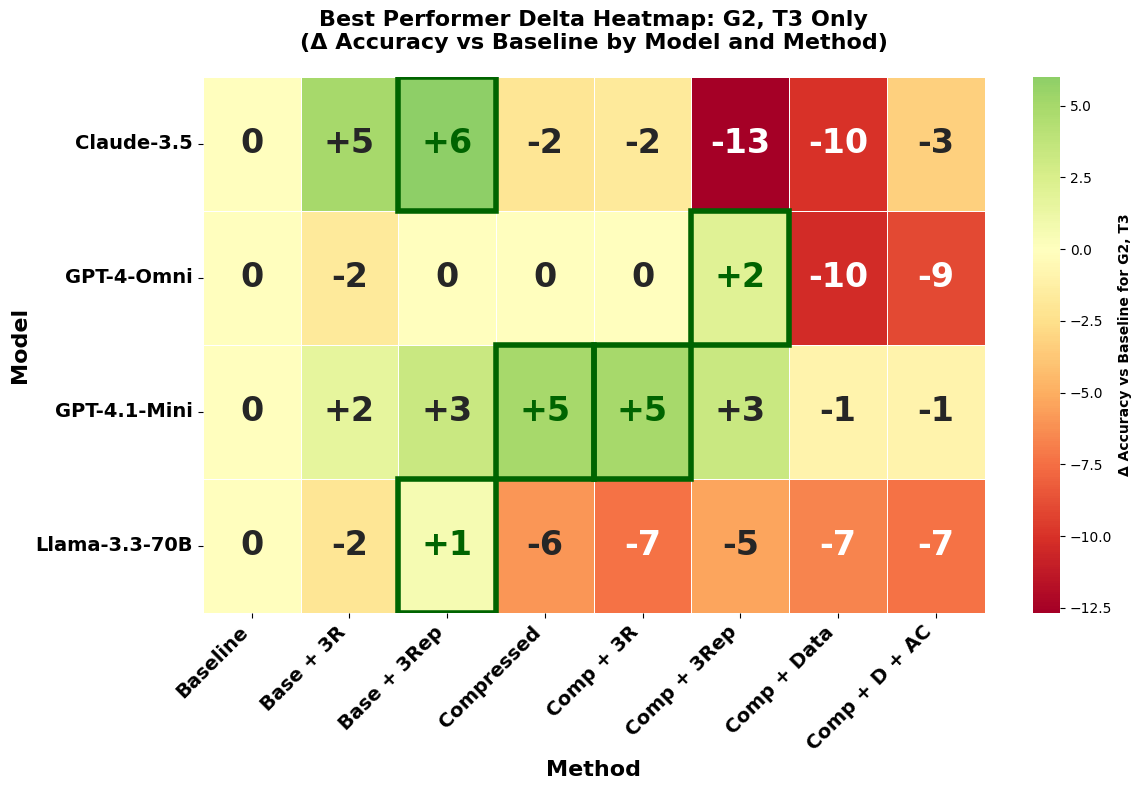}
  \caption{Accuracy deltas ($\Delta$) relative to baseline on TAT-QA with n-gram size 2 and Top-3 abbreviation. Claude-3.5 and GPT-4.1-Mini show improvements up to +6 and +5 points, while GPT-4-Omni and Llama-3.3-70B remain mostly stable or decline under compression.}
  \label{fig:heattatqa}
\end{figure}

\begin{figure}[h]
  \centering
  \includegraphics[width=0.5\textwidth]{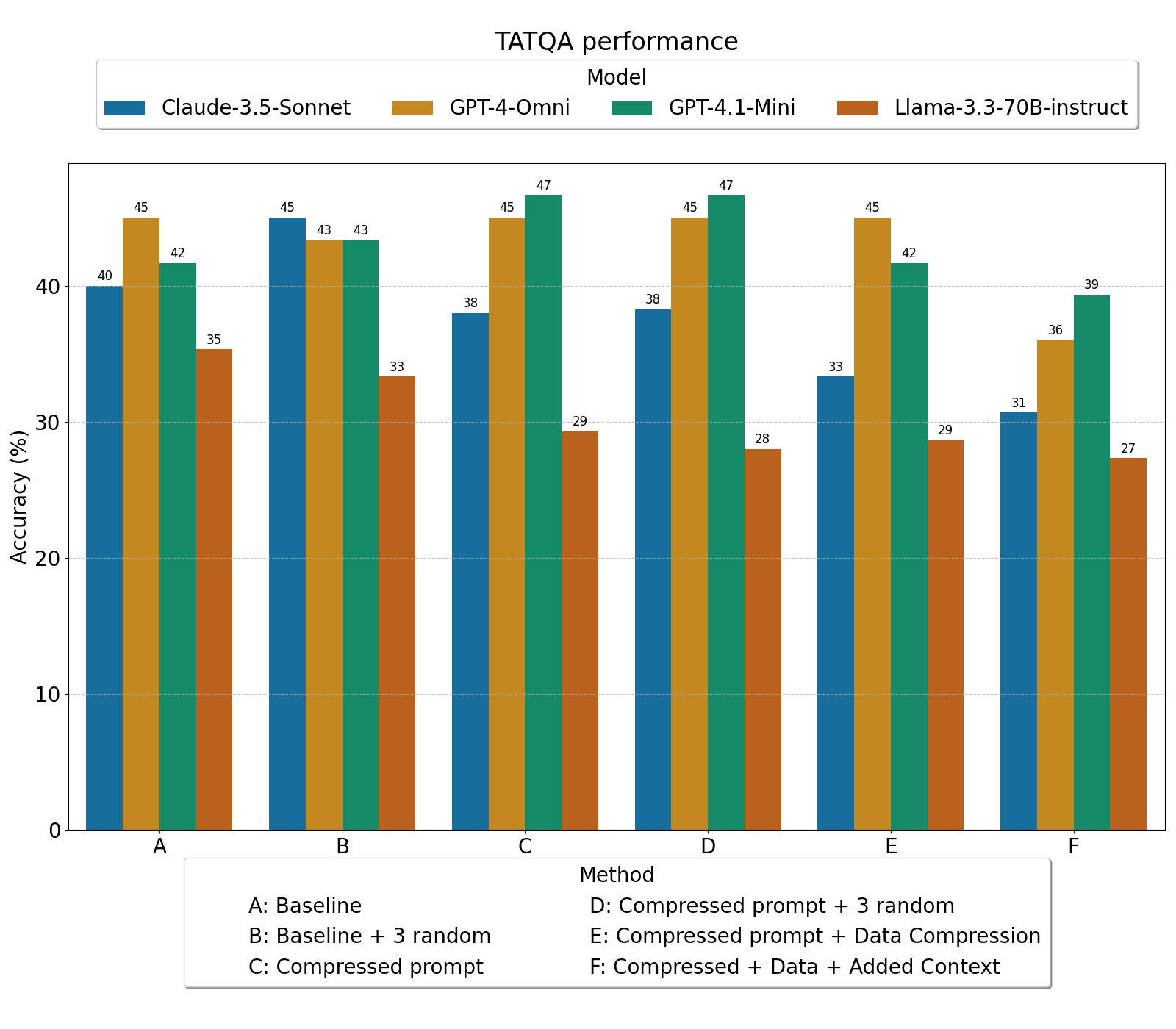}
  \caption{Accuracy on TAT-QA with baseline and compression variants using three random exemplars. GPT-4.1-Mini benefits from compression (+5\%), while Claude-3.5 and GPT-4-Omni maintain stable performance, and Llama-3.3-70B shows declines under most compressed settings.}
  \label{fig:acctatqa}
\end{figure}

We evaluate CompactPrompt on the TAT-QA benchmark, which requires cross-referencing structured tables with accompanying narrative passages. Figure \ref{fig:acctatqa} shows absolute model accuracies across prompting methods, while Figure \ref{fig:heattatqa} presents accuracy deltas relative to baseline under the best-performing hyperparameter setting $(T=3, G=2)$. \\

\textbf{Baseline vs. Compression:} As shown in Figure \ref{fig:acctatqa}, uncompressed prompts (Method A) yield accuracies between 33–45\% depending on the model. Applying CompactPrompt’s compression (Method C) maintains this baseline performance and, in several cases, improves it. The heatmap in Figure \ref{fig:heattatqa} highlights that Claude 3.5 Sonnet (+6) and GPT-4.1-Mini (+5) achieve the largest gains, GPT-4-Omni remains stable, and Llama-3.3-70B shows a modest improvement (+1). Together, the two plots confirm that CompactPrompt reduces prompt size without sacrificing accuracy, while often delivering a measurable boost. 

\textbf{Effect of Exemplars}

When compression is combined with exemplar-based prompting, Figures \ref{fig:heattatqa} show effects on adding the random exampels from the TATQA dataset. Adding three representative examples (Method C and E) sustains the gains for Claude 3.5 Sonnet and GPT-4.1-Mini (both +5) while keeping GPT-4-Omni near baseline. In contrast, adding random exemplars (Method B or D-random) and also shown in \ref{fig:acctatqa} produces less consistent improvements, reinforcing the importance of careful example selection. 

\textbf{Data Compression and Added Context}

Figures \ref{fig:acctatqa} further illustrate that integrating document-level data compression (Method E) or reintroducing the abbreviation dictionary as added context (Method F) yields mixed outcomes. For GPT-4.1-Mini, performance remains same level baseline, but other models exhibit small declines. This suggests that while textual abbreviation provides a reliable efficiency–fidelity trade-off, tabular data compression must be applied selectively to avoid disrupting cross-referential reasoning. 

\textbf{Cross-model insights:}

\begin{itemize}
    \item Claude 3.5 Sonnet is the most robust across settings, consistently outperforming its baseline under compression in both Figures \ref{fig:heattatqa} and \ref{fig:acctatqa}. 

\item GPT-4.1-Mini also shows dependable improvements, particularly with representative exemplars. 

\item GPT-4-Omni maintains stable accuracy across all settings, showing that CompactPrompt introduces no regressions. 

\item Llama-3.3-70B improves slightly under light compression but is more sensitive to aggressive configurations, as reflected in the heatmap’s negative deltas. 
\end{itemize}

\textbf{Summary of findings in TATQA: }
 
Across both Figures \ref{fig:heattatqa} and \ref{fig:acctatqa}, CompactPrompt achieves up to +6 accuracy points while cutting token usage by more than half. The key outcome is that model performance is preserved under compression, fulfilling the primary goal of efficiency without loss. The consistent gains for Claude 3.5 Sonnet and GPT-4.1-Mini demonstrate that compression can even enhance reasoning quality—a welcome bonus beyond cost reduction. 

\begin{figure}[h]
  \centering
  \includegraphics[width=0.5\textwidth]{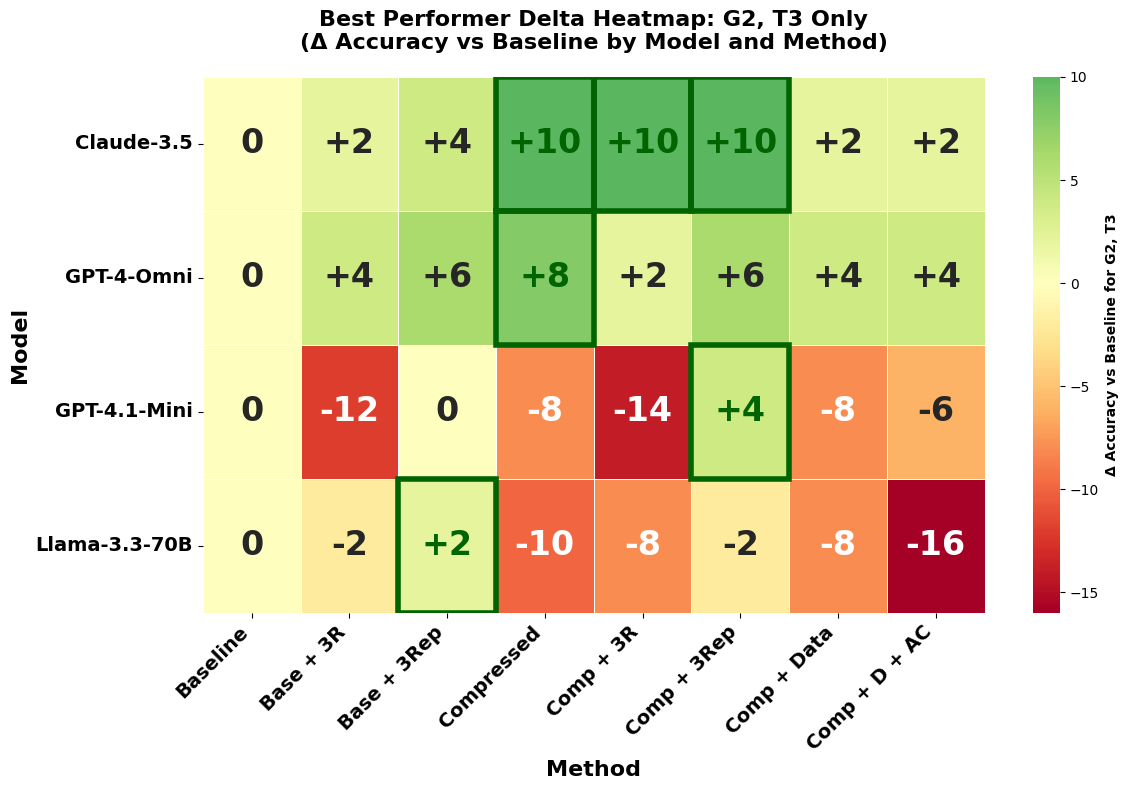}
  \caption{Accuracy deltas ($\Delta$) relative to baseline on FinQA with n-gram size 2 and Top-3 abbreviation. Claude-3.5 and GPT-4-Omni achieve gains up to +10 and +8 points, while GPT-4.1-Mini and Llama-3.3-70B experience notable drops.}
  \label{fig:heatfinqa}
\end{figure}

\begin{figure}[h]
  \centering
  \includegraphics[width=0.5\textwidth]{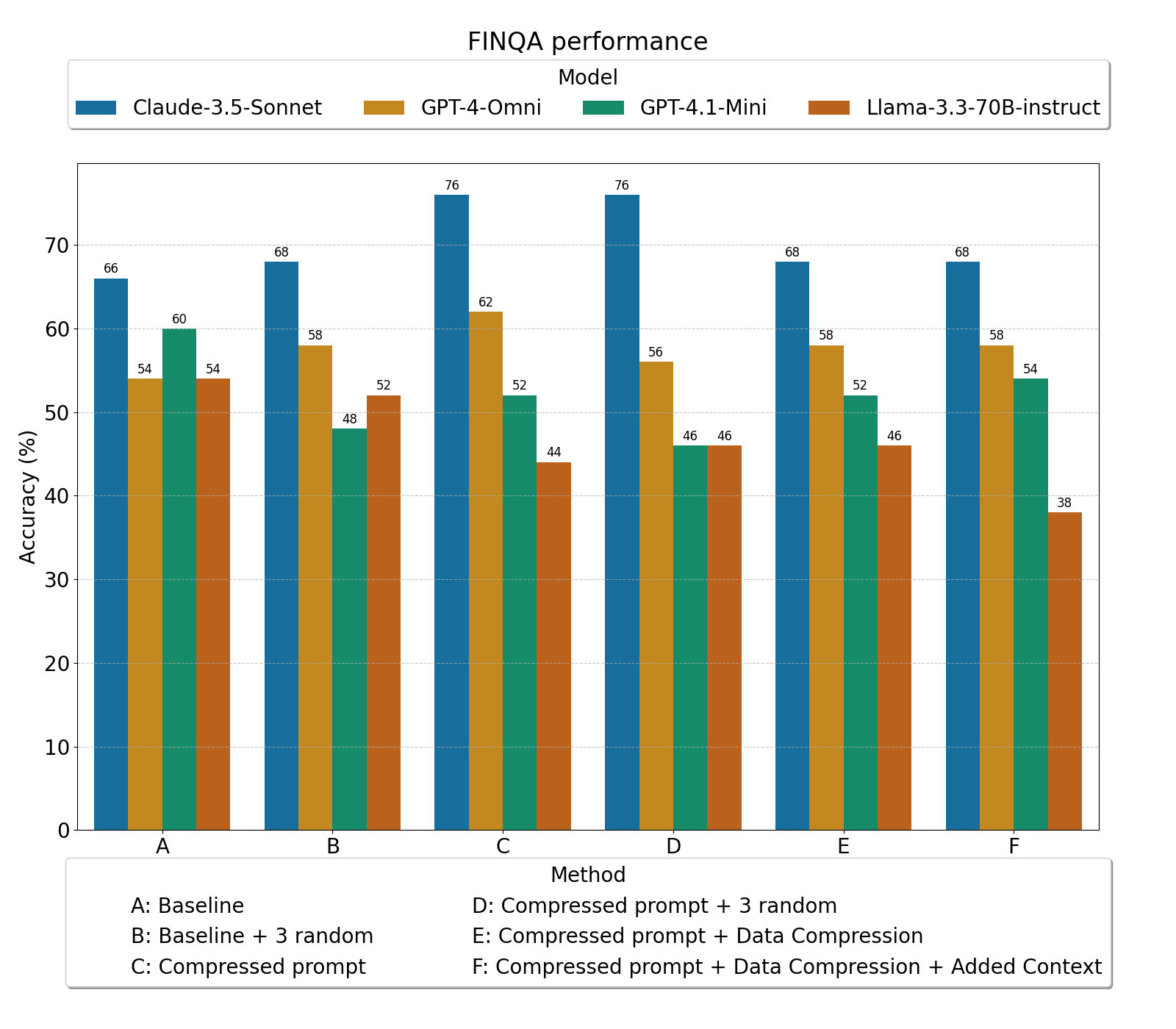}
  \caption{Accuracy on FinQA with baseline and compression variants using three random exemplars. Claude-3.5 achieves the highest performance (76\%) under compressed prompt settings, while GPT-4.1-Mini and Llama-3.3-70B show reduced stability under compression.}
  \label{fig:accFinQA_random}
\end{figure}

\begin{figure}[h]
  \centering
  \includegraphics[width=0.5\textwidth]{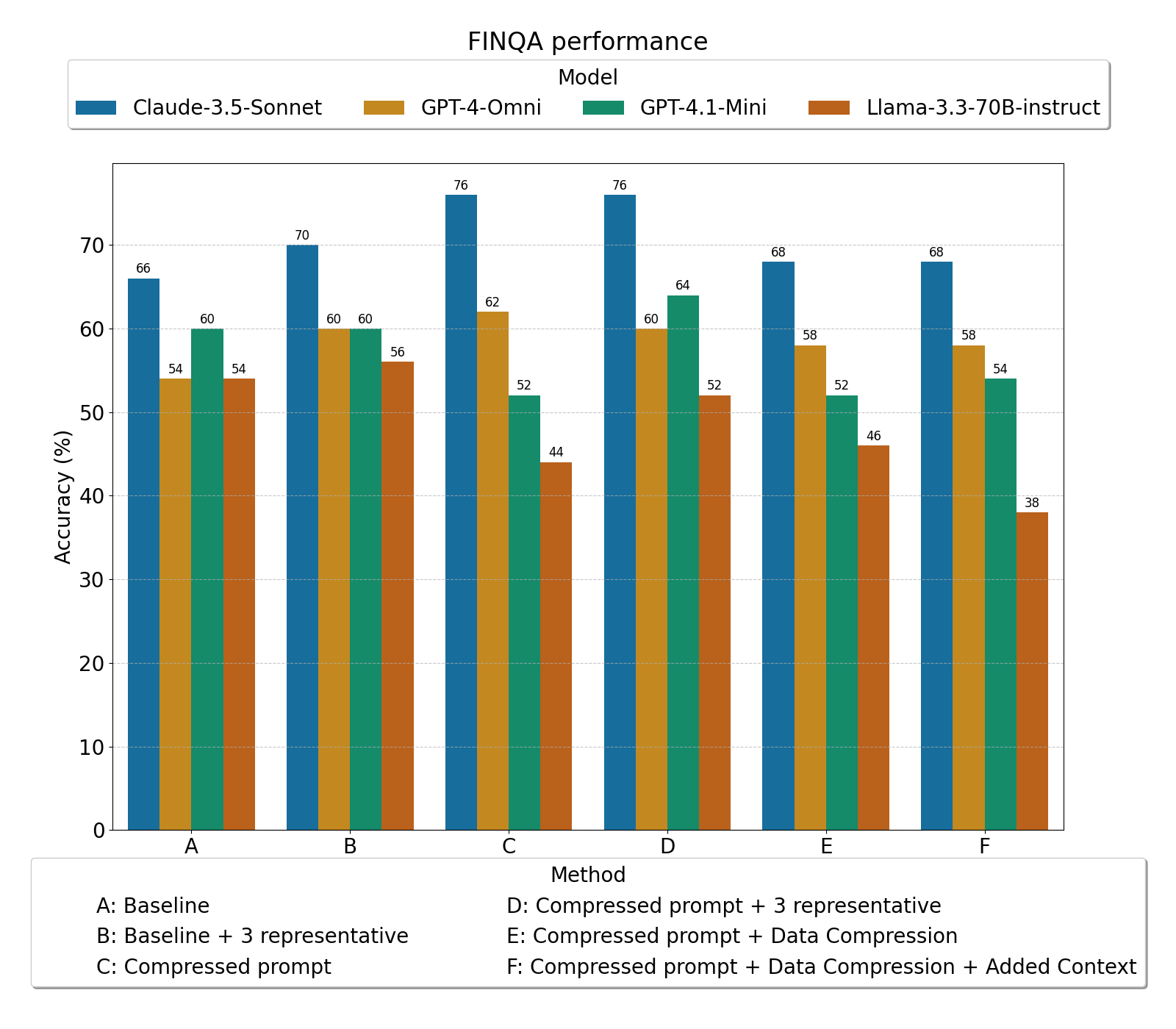}
  \caption{Accuracy on FinQA with baseline and compression variants using three representative exemplars. Representative sampling yields stronger gains for GPT-4.1-Mini (+5\% over baseline) and more consistent improvements across models compared to random exemplars.}
  \label{fig:accFinQA_representative}
\end{figure}

We further evaluate CompactPrompt on Fin-QA, a benchmark emphasizing numerical reasoning over financial text. Figures \ref{fig:accFinQA_random} and \ref{fig:accFinQA_representative} report absolute accuracies with random versus representative exemplars, while Figure \ref{fig:heatfinqa} shows deltas relative to baseline under the best-performing hyperparameter setting $(G=2, T=3)$. 

\textbf{Baseline vs. Compression}
 Figure \ref{fig:accFinQA_random} shows baseline accuracies ranging from 54–66\% across models. Compressed prompts (Method C) maintain these levels for most models and even surpass them in some cases: Claude 3.5 Sonnet improves from 66\% to 76\% (+10), while GPT-4-Omni also achieves notable gains (+6 to +8 as highlighted in Figure \ref{fig:heatfinqa}). These improvements demonstrate that CompactPrompt can preserve numerical reasoning accuracy while reducing token counts, and in certain cases enhance performance. 

\textbf{Random vs. Representative Exemplars}
 A clear distinction emerges when comparing Figures \ref{fig:accFinQA_random} and \ref{fig:accFinQA_representative}. With random exemplars, results are less consistent: GPT-4.1-Mini and Llama-3.3-70B occasionally show declines, indicating sensitivity to exemplar choice. However, with representative exemplars (Figure \ref{fig:accFinQA_representative}), performance stabilizes across models. Claude 3.5 Sonnet sustains its +10 point gain, GPT-4-Omni improves by up to +8, and even GPT-4.1-Mini recovers from its earlier declines to exceed baseline in selected configurations. This highlights the importance of exemplar quality: representative examples yield reliable improvements under compression, while random examples can introduce noise. 

\textbf{Data Compression and Added Context} 
 As with TAT-QA, integrating data-level compression (Method E) or appending abbreviation dictionaries (Method F) yields mixed results. In Figure \ref{fig:heatfinqa}, Claude 3.5 Sonnet maintains +10 across compressed settings, and GPT-4-Omni shows stable gains. By contrast, GPT-4.1-Mini and Llama-3.3-70B are more sensitive, sometimes registering modest declines. These outcomes suggest that while CompactPrompt consistently preserves baseline accuracy, aggressive data compression should be tuned carefully depending on the model. 

\textbf{Cross-model insights}

\begin{itemize}
\item Claude 3.5 Sonnet benefits most, achieving sustained improvements of +10 points in compressed conditions (Figures \ref{fig:accFinQA_random} and \ref{fig:accFinQA_representative}). 

\item GPT-4-Omni also gains consistently (+6 to +8), showing strong robustness. 

\item GPT-4.1-Mini is more sensitive to exemplar choice: it declines with random examples but recovers when representative ones are used. 

\item Llama-3.3-70B shows smaller gains (+2 in some settings) and is more affected by aggressive compression. 
\end{itemize}

\textbf{Summary of findings in Fin-QA: }
Across Figures \ref{fig:heatfinqa} – \ref{fig:accFinQA_representative}, FinQA experiments confirm that CompactPrompt achieves substantial token savings without accuracy loss, fulfilling the primary goal of compression. Moreover, for Claude 3.5 Sonnet and GPT-4-Omni, compression paired with representative examples yields clear accuracy improvements (+6 to +10 points), showing that efficiency and reasoning quality can be advanced together when hyperparameters and exemplars are tuned appropriately. \\

\textbf{Overall Insights: }
Across both TAT-QA and Fin-QA, the results demonstrate that CompactPrompt reliably reduces token usage by more than half while preserving baseline accuracy—a crucial outcome for practical deployment. The ablation study highlights that careful tuning of abbreviation hyperparameters (G=2,T=3) is key to striking the right balance between efficiency and fidelity. Representative exemplars further amplify these benefits, stabilizing performance and enabling consistent gains across models. While Claude 3.5 Sonnet and GPT-4-1-Mini show the strongest improvements, GPT-4-Omni remains robust under all settings, and even Llama-3.3-70B benefits under lighter compression. Taken together, these findings confirm that CompactPrompt can be deployed with confidence: the primary goal of efficiency is achieved without loss, and in many cases, accuracy is enhanced—a positive signal for integrating compression into high-stakes financial workflows.

\section{Conclusion}

CompactPrompt demonstrates that a training-free, hard-compression approach can substantially reduce prompt length, achieving up to 60\%  token savings on TAT-QA and FinQA while maintaining, and in some cases improving, QA accuracy. Representative exemplar selection further enhances performance by capturing the underlying characteristics of the dataset more faithfully than random exemplars. We also show that high embedding-based semantic similarity ($\geq$ 0.92) generally indicates safe compression, though moderate similarity drops do not necessarily harm downstream performance. Interestingly, prompts compressed with Claude 3.5 Sonnet exhibited slightly lower semantic embedding similarity (0.941) than those from GPT-4.1-Mini (0.952), yet performed better on TAT-QA—underscoring that semantic metrics alone cannot guarantee task-level fidelity. We therefore advocate a multi-pronged evaluation strategy combining embedding-based thresholds with human review, LLM-as-a-judge assessments, and direct task accuracy checks to ensure that efficiency gains do not come at the expense of quality.

Global financial institutions have proven to be among the earliest adopters of generative AI. As banks pursue democratization of LLM-based systems, shareholders rightly expect both return on investment and strict adherence to regulatory standards around privacy. Prompt compression directly supports both goals: lowering costs by reducing token usage and simultaneously offering a pathway to limit inadvertent disclosure of sensitive client or firm data.

\section{Future Work}
 Looking ahead, several promising directions can extend the utility of CompactPrompt:

\begin{itemize}
\item \textbf{Adaptive compression.} Future versions could dynamically adjust the compression budget based on task complexity, context length, or real-time performance signals.
\item \textbf{Privacy-aware compression.} Assigning higher weights to sensitive fields (\emph{e.g.}, identifiers, account numbers, social security numbers) would allow compression to function as a lightweight privacy filter, removing high-risk tokens before data reaches an LLM.
\item \textbf{Integration with enterprise pipelines.} Embedding CompactPrompt into financial workflows such as regulatory reporting, automated compliance, and portfolio monitoring could help standardize cost-efficient deployment at scale.
\item \textbf{Beyond text and tables.} Extending compression techniques to multimodal contexts (\emph{e.g.}, charts, PDF documents, or time-series feeds) would broaden applicability to real-world financial analysis tasks.
\item \textbf{Reduced reliance on RAG.} While retrieval-augmented generation (RAG) remains powerful, compression could enable larger proprietary documents to be directly incorporated into prompts, offering faster inference and simpler deployment when RAG infrastructure is costly to maintain.
\end{itemize}

CompactPrompt is proprietary software owned by The Bank of New York (BNY). In accordance with company policies and intellectual property regulations, we are unable to make the tool publicly available. Access to the tool is restricted to internal use and authorized personnel only hence external distribution or open-source release is not permitted.

\bibliographystyle{ACM-Reference-Format}
\bibliography{main}

\appendix
\clearpage      

\onecolumn 
\section{Appendix}

\small
\begin{longtable}{ccrrrr}
\caption{TAT-QA Accuracy (\%) for different models, methods, Top-N and n-gram under Baseline setting}
\label{tab:TATQA-results}\\

\toprule
Top N/N-gram & Method & Claude-3.5-Sonnet & GPT-4-Omni & GPT-4.1-Mini & Llama-3.3-70B-instruct \\
\midrule
\endfirsthead

\toprule
\multicolumn{6}{l}{\textit{Table \thetable{} (continued)}}\\
\midrule
Top N/N-gram & Method & Claude-3.5-Sonnet & GPT-4-Omni & GPT-4.1-Mini & Llama-3.3-70B-instruct \\
\midrule
\endhead

\midrule
\multicolumn{6}{r}{\textit{Continued on next page}}\\
\bottomrule
\endfoot

\bottomrule
\endlastfoot

3/2 & Baseline & 40 & 45 & 42 & 35 \\
3/2 & Baseline + 3 random & 45 & 43    & 43    & 33  \\
3/2 & Baseline + 3 representative & 46 & 45 & 45 & 36 \\

\end{longtable}

 \small 
\begin{longtable}{ccrrrr}
\caption{TAT-QA Accuracy (\%) for different models, methods, Top-N and n-gram under compressed setting}
\label{tab:TATQA-results}\\

  \toprule
Top N/N-gram & Method & Claude-3.5-Sonnet & GPT-4-Omni & GPT-4.1-Mini & Llama-3.3-70B-instruct  \\
\midrule
\endfirsthead

\toprule
\multicolumn{6}{l}{\textit{Table \thetable{} (continued)}}\\
\midrule
Top N / N-gram & Method & Claude-3.5-Sonnet & GPT-4-Omni & GPT-4.1-Mini & Llama-3.3-70B-instruct \\
\midrule
\endhead

\midrule
\multicolumn{6}{r}{\textit{Continued on next page}}\\
\bottomrule
\endfoot

\bottomrule
\endlastfoot
    
3/2 & Compressed Prompt & 38 & 45 & 47 & 29 \\
3/2 & Compressed + 3 random & 38 & 45 & 47 & 28 \\
3/2 & Compressed + 3 representative & 27 & 47 & 45 & 30 \\
\end{longtable}

\small 
\begin{longtable}{crrrr}
\caption{TAT-QA Accuracy (\%) for different models, methods, Top-N and n-gram under Compressed + Data}
\label{tab:TATQA-results}\\

  \toprule
Top N/N-gram & Claude-3.5-Sonnet & GPT-4-Omni & GPT-4.1-Mini & Llama-3.3-70B-instruct  \\
\midrule
\endfirsthead

\toprule
\multicolumn{5}{l}{\textit{Table \thetable{} (continued)}}\\
\midrule
Top N / N-gram & Claude-3.5-Sonnet & GPT-4-Omni & GPT-4.1-Mini & Llama-3.3-70B-instruct \\
\midrule
\endhead

\midrule
\multicolumn{5}{r}{\textit{Continued on next page}}\\
\bottomrule
\endfoot

\bottomrule
\endlastfoot
 
2/2 & 31 & 34 & 39 & 29 \\
2/3 & 33 & 45 & 42 & 29 \\
2/4 & 31 & 35 & 43 & 31 \\
3/2 & 30 & 35 & 41 & 29 \\
3/3 & 34 & 35 & 39 & 30 \\
3/4 & 31 & 37 & 39 & 32 \\
4/2 & 31 & 35 & 42 & 31 \\
4/3 & 35 & 35 & 43 & 29 \\
4/4 & 33 & 36 & 42& 30 \\
\end{longtable}

\small 
\begin{longtable}{crrrr}
\caption{TAT-QA Accuracy (\%) for different models, methods, Top-N and n-gram under Compressed + Data + Added Context}
\label{tab:TATQA-results}\\

  \toprule
Top N/N-gram & Claude-3.5-Sonnet & GPT-4-Omni & GPT-4.1-Mini & Llama-3.3-70B-instruct  \\
\midrule
\endfirsthead

\toprule
\multicolumn{5}{l}{\textit{Table \thetable{} (continued)}}\\
\midrule
Top N / N-gram & Claude-3.5-Sonnet & GPT-4-Omni & GPT-4.1-Mini & Llama-3.3-70B-instruct \\
\midrule
\endhead

\midrule
\multicolumn{5}{r}{\textit{Continued on next page}}\\
\bottomrule
\endfoot

\bottomrule
\endlastfoot
    
2/2 & 38 & 36 & 42    & 27    \\
2/3 & 31 & 36 & 39    & 27    \\
2/4 & 29 & 35 & 41    & 28   \\
3/2 & 37 & 36 & 41    & 28   \\
3/3 & 33 & 37 & 41 & 27 \\
3/4 & 29 & 37 & 40 & 29 \\
4/2 & 37 & 35 & 41 & 30 \\
4/3 & 32 & 33 & 42 & 29 \\
4/4 & 30 & 33 & 41 & 29 \\
\end{longtable}

\newpage
\small 
\begin{longtable}{ccrrrr}
\caption{FinQA Accuracy (\%) for different models, methods, Top-N and n-gram for Baseline}
\label{tab:TATQA-results}\\

  \toprule
Top N/N-gram & Method & Claude-3.5-Sonnet & GPT-4-Omni & GPT-4.1-Mini & Llama-3.3-70B-instruct \\
 \\
\midrule
\endfirsthead

\toprule
\multicolumn{6}{l}{\textit{Table \thetable{} (continued)}}\\
\midrule
Top N/N-gram & Method & Claude-3.5-Sonnet & GPT-4-Omni & GPT-4.1-Mini & Llama-3.3-70B-instruct \\
 \\
\midrule
\endhead

\midrule
\multicolumn{6}{r}{\textit{Continued on next page}}\\
\bottomrule
\endfoot
\bottomrule
\endlastfoot

3/2 & Baseline & 66 & 54 & 60 & 54 \\
3/2 & Baseline + 3 random & 68 & 58 & 48 & 52 \\
3/2 & Baseline + 3 representative & 70 & 60 & 60 & 56 \\

\end{longtable}

 \small 
\begin{longtable}{ccrrrr}
\caption{FinQA Accuracy (\%) for different models, methods, Top-N and n-gram under compressed setting}
\label{tab:TATQA-results}\\

  \toprule
Top N/N-gram & Method & Claude-3.5-Sonnet & GPT-4-Omni & GPT-4.1-Mini & Llama-3.3-70B-instruct  \\
\midrule
\endfirsthead

\toprule
\multicolumn{6}{l}{\textit{Table \thetable{} (continued)}}\\
\midrule
Top N / N-gram & Method & Claude-3.5-Sonnet & GPT-4-Omni & GPT-4.1-Mini & Llama-3.3-70B-instruct \\
\midrule
\endhead

\midrule
\multicolumn{6}{r}{\textit{Continued on next page}}\\
\bottomrule
\endfoot

\bottomrule
\endlastfoot
    
3/2 & Compressed Prompt & 76 & 62 & 52 & 44 \\
3/2 & Compressed + 3 random & 76 & 56 & 46 & 46 \\
3/2 & Compressed + 3 representative & 76 & 60 & 64 & 52 \\
\end{longtable}

\small 
\begin{longtable}{ccrrr}
\caption{FinQA Accuracy (\%) for different models, methods, Top-N and n-gram for Compressed + Data}
\label{tab:TATQA-results}\\

  \toprule
Top N/N-gram & Claude-3.5-Sonnet & GPT-4-Omni & GPT-4.1-Mini & Llama-3.3-70B-instruct \\
 \\
\midrule
\endfirsthead

\toprule
\multicolumn{5}{l}{\textit{Table \thetable{} (continued)}}\\
\midrule
Top N/N-gram & Claude-3.5-Sonnet & GPT-4-Omni & GPT-4.1-Mini & Llama-3.3-70B-instruct \\
 \\
\midrule
\endhead

\midrule
\multicolumn{5}{r}{\textit{Continued on next page}}\\
\bottomrule
\endfoot
\bottomrule
\endlastfoot

2/2 & 74&54&56&44 \\
2/3 & 74&38&60&42 \\
2/4 & 70&36&62&42 \\
3/2 & 68&58&52&46 \\
3/3 & 74&56&60&42 \\
3/4 & 74&44&58&44 \\
4/2 &72&48&58&44 \\
4/3 &74&48&56&42 \\
4/4 &78&54&58&48 \\
5/2&74&44&62&46 \\
5/3&70&48&58&44 \\
5/4 &76&58&56&42
 \end{longtable}

\small 
\begin{longtable}{crrrr}
\caption{FinQA Accuracy (\%) for different models, methods, Top-N and n-gram for Compressed + Data + Added Context }
\label{tab:TATQA-results}\\

\toprule
Top N/N-gram & Claude-3.5-Sonnet & GPT-4-Omni & GPT-4.1-Mini & Llama-3.3-70B-instruct \\
 \\
\midrule
\endfirsthead

\toprule
\multicolumn{5}{l}{\textit{Table \thetable{} (continued)}}\\
\midrule
Top N/N-gram & Claude-3.5-Sonnet & GPT-4-Omni & GPT-4.1-Mini & Llama-3.3-70B-instruct \\
 \\
\midrule
\endhead

\midrule
\multicolumn{5}{r}{\textit{Continued on next page}}\\
\bottomrule
\endfoot
\bottomrule
\endlastfoot

2/2 & 72&42&56&46 \\
2/3 & 70&46&64&46\\
2/4 & 70&44&58&42\\
3/2 &68&58&54&38\\
3/3 & 74&46&58&42\\
3/4 &74&50&56&38\\
4/2 &64&44&54&44\\
4/3&72&40&58&42\\
4/4&66&54&54&40\\
5/2&68&44&56&38\\
5/3&70&22&60&38\\
5/4&66&60&58&42\\

\end{longtable}

\end{document}